\documentclass[twocolumn]{autart}    

\usepackage{svg}

\usepackage{graphicx,import}

\usepackage{bm}
\usepackage{amssymb}
\usepackage{amsmath}

\usepackage{makecell}

\usepackage{bigints}  

\usepackage{array}

\usepackage{xcolor}

\begin{document}

\begin{frontmatter}

\title{An accurate and efficient approach to probabilistic conflict prediction} 



\author {Christian E. Roelofse}\ead{17003342@sun.ac.za},    
\author{Corn\'e E. van Daalen}\ead{cvdaalen@sun.ac.za},               

\address[South Africa]{Department of Electrical and Electronic Engineering, Stellenbosch University, Stellenbosch 7600}             


\vspace{-0.6cm} 
\begin{abstract}                          
Conflict prediction is a vital component of path planning for autonomous vehicles. Prediction methods must be accurate for reliable navigation, but also computationally efficient to enable online path planning. Efficient prediction methods are especially crucial when testing large sets of candidate trajectories. We present a prediction method that has the same accuracy as existing methods, but up to an order of magnitude faster. This is achieved by rewriting the conflict prediction problem in terms of the first-passage time distribution using a dimension-reduction transform. First-passage time distributions are analytically derived for a subset of Gaussian processes describing vehicle motion. The proposed method is applicable to 2-D stochastic processes where the mean can be approximated by line segments, and the conflict boundary can be approximated by piece-wise straight lines. The proposed method was tested in simulation and compared to two probability flow methods, as well as a recent instantaneous conflict probability method. The results demonstrate a significant decrease of computation time. 
\end{abstract}

\end{frontmatter}
\vspace{-0.3cm} 
\section{Introduction}
Autonomous navigation is the process whereby a vehicle traverses an environment without human intervention. This process can be construed as a combination of mapping, localisation, and motion planning. The vehicle localises itself in a map of its environment, and should plan a safe trajectory through the map. However, this process is complicated by the influence of uncertainty: unknown disturbances, measurement error, and prediction uncertainty. Reliable conflict (or collision) prediction  is therefore an essential component of a motion planning subsystem.

A motion planner typically generates a large set of candidate trajectories for the vehicle \cite{LaValle}, which needs to be inspected for conflict to ensure safe navigation. Conflict predictions need to be accurate to be reliable for the purpose of navigation. However, since the computational cost associated with conflict predication accumulates for each path tested, it needs to be computationally efficient to allow for online autonomous navigation. Conflict prediction methods can be divided into \emph{deterministic} and \emph{probabilistic} methods \cite{van2009fastMyP28}. Deterministic methods produce a binary result and do not account for uncertainty \cite{Yang_survy_2000MyP18}. In contrast, probabilistic methods calculate the probability of conflict, which accounts for uncertainty and allow for more sophisticated risk reasoning \cite{jansson2008framework}. The characteristics of two deterministic methods (nominal and worst-case) and the probabilistic method are compared and summarised in Table \ref{myTab_1} \cite{Yang_survy_2000MyP18}. Probabilistic methods are therefore preferable \cite{van2009fastMyP28} due to their ability to handle uncertainty; however, they are usually computationally expensive.

Probabilistic conflict prediction is typically solved using either a sampling-based, analytical or numerical approach. Sampling-based approaches (Monte Carlo methods) allow for nonparametric distributions and nonlinear motion models \cite{jansson2008framework}. However, such methods tend to be expensive, even when using techniques such as importance sampling \cite{chr2010improvedMyP3}. Analytical approaches can result in a closed-form solution, typically made possible by strict modelling assumptions, or certain kinds of vehicle encounter conditions \cite{Erzberger_paielli_1997MyP21}. These limitations are necessary to produce a closed-form expression, but are often too restrictive to be used for most applications. Numerical approaches are based on less strict assumptions, and use numerical techniques where needed \cite{NumeCiteArticle}. In general, the assumptions made by numerical approaches are not as restrictive as those made by analytical approaches. Numerical approaches also take advantage of specific choices of motion and noise models, unlike sampling methods. Therefore, the cost of numerical techniques is not typically as expensive as sampling. Three probabilistic solution approaches are summarised in Table \ref{myTab_1}.

\begin{center}
\begin{table}[ht]
    \centering
    \caption{Qualitative summary of conflict prediction methods and probabilistic solution approaches. Descriptions are meant to be interpreted relative to the descriptions of competitors.}\label{myTab_1} \vspace{0.1cm}
    \begin{tabular}{| m{2.3cm} | m{2.3cm} | m{2.3cm}|}
  \Xhline{3\arrayrulewidth} 

    \multicolumn{3}{|c|}{\textbf{Prediction methods}} \\ 
    \Xhline{3\arrayrulewidth} 

      \centering Nominal  & \centering Worst-case & \centering{ Probabilistic } \cr \hline   
       
      \centering Lower cost & \centering Moderate cost & \centering Higher cost \cr
      \centering  Disregards uncertainty & \centering Deterministically accounts for uncertainty & \centering Accounts for uncertainty \cr
       \centering Uses nominal path & \centering Uses worst-case path envelope & \centering Uses sample paths \cr
       \centering Optimistic & \centering Pessimistic & \centering More realistic \cr 
       
      \Xhline{3\arrayrulewidth} 
      
    \multicolumn{3}{|c|}{\textbf{Probabilistic solution approaches}} \\  \Xhline{3\arrayrulewidth}  
      \centering Analytical  & \centering Sampling & \centering{ Numerical } \cr \hline   
      
      \centering Restrictive assumptions& \centering Minimal assumptions& \centering Moderate assumptions \cr
      \centering Lower cost& \centering Higher cost& \centering Moderate cost \cr
      \centering Accuracy varies\footnotemark[1]& \centering Accurate & \centering Accuracy varies\footnotemark[1] \cr
      \hline     
    \end{tabular}
\end{table}
\end{center} \footnotetext[1]{Varies depending on the degree to which the assumptions conform to the application.}
\vspace{-0.7cm}

Two examples of probabilistic prediction methods, which both use a numerical solution, are probability flow (PF) and instantaneous conflict probability (ICP). The PF method of van Daalen and Jones \cite{van2009fastMyP28} is a fast, upper bound method that numerically integrates the rate of probability mass flowing over the conflict boundary (CB). Park and Kim devised a new method that separates PF into drift and diffusion components, which are likewise integrated over the CB \cite{flow_As_2017MyP22}. ICP methods calculate the instantaneous overlap of probability mass and the conflict region over time, which is used to calculate conflict probability. The rectangular region approximation method of Pour et al. \cite{pour2019probability} is a recent example of an ICP method. 

The focus of this article is the derivation of an efficient, probabilistic prediction method. The envisioned application is general autonomous navigation, which requires efficient conflict prediction to avoid a computational bottleneck in the motion planner. However, the proposed method can also be used in other contexts, such as advisory systems for piloted vehicles. The derivation presented uses a novel application of the first-passage time (FPT) metric with a dimension-reduction approach, applied to a subset of Gaussian processes. The proposed method is a numerical approach where the conflict prediction problem is rewritten in terms of the FPT distribution, which makes use of analytically derived FPT expressions. The dimension-reduction is achieved by using line segments as conflict boundaries, which can be used to construct an arbitrary boundary shape. The proposed method is applicable to 2-D scenarios where the vehicle's mean motion can be approximated by line segments, and the conflict boundary can be approximated by piece-wise line segments. The state distribution of the vehicle also needs to drift or diffuse in the direction of the conflict boundary.

The rest of the article is organised as follows: Section 2 outlines the problem formulation, then Section 3 gives an overview of the first-passage time problem. Sections 4 and 5 presents the solution formulation and derivation. Finally, Section 6 ends with a simulation comparison, highlighting the performance of the proposed method.

\section{Problem Formulation}

Before the solution development is presented, some problem-specific assumptions are stated along with a definition of the problem. The research presented is intended to function in a \emph{decentralised}, non-cooperative context \cite{Dimarogonas2002MyP4}. All vehicles are assumed to behave \emph{independently}, which means that the predicted states of other vehicles are the same for all possible planned trajectories, and that the other vehicles do not react to the chosen trajectory, or subsequent manoeuvring. The vehicle and environment are assumed to be \emph{state observable} \cite{chr2010improvedMyP3}: the states can be estimated from the available information. Vehicle state estimates are chosen to be Gaussian distributed, which is typical for problem definitions involving vehicle motion \cite{van2009fastMyP28,Erzberger_paielli_1997MyP21,jones2006ArticleMyProp11}, and the noise that models the influence of disturbances is white, uncorrelated and zero-mean. The noise approximates the effects of disturbances such as wind on the vehicle.

This research considers \emph{conflict} prediction: predicting the violation of a user-defined, minimum separation, in contrast to \emph{collision} prediction \cite{van2009fastMyP28}. This keep-out region defines a conflict region set $\bm{\mathcal{D}}$ in the environment. Nevertheless, the solution presented in this article can be used for collision prediction, since collision by definition entails conflict. The desired output of conflict prediction is the probability $\mathbb{P}_C$ that a vehicle will enter a conflict region $\bm{\mathcal{D}}$ over a finite prediction period $t_H$. Starting from $t=0$, the conflict outcomes can be expressed as a set \cite{van2009fastMyP28}
\begin{align}
\label{set_A}
A_{t_H}  =  \{\omega \in \Omega \ : \ \exists\ t_C \in [0,t_H], \ \mathbf{R}(t_C, \omega)  \in \bm{\mathcal{D}} \},
\end{align}
where $\omega \in \Omega$ represents a stochastic trajectory outcome and $\mathbf{R}(t, \omega)$ denotes the position state of a vehicle. Therefore, the probability of conflict can be expressed as
\begin{align}
\label{prob_calc_basic}
\mathbb{P}_C(t_H)  =  P[A_{t_H}].
\end{align}
The vehicle is assumed to be a point mass with Gaussian state uncertainty. The conflict region $\bm{\mathcal{D}}$ is assumed to have no uncertainty. The problem description and assumptions can be related to a more realistic problem definition by making appropriate transforms; the uncertainty of both the vehicle position and the conflict region position can be combined into a single equivalent source of uncertainty, centred on the vehicle. The volume of the vehicle can be used to define a keep-out region centred on the vehicle. As such, the keep-out region of the vehicle can likewise be combined with that of the conflict region in the environment. These transforms result in an equivalent problem description: calculating the conflict probability of a point mass with uncertain state, with respect to a certain, combined conflict region \cite{van2010conflictMyP27}. This region can be determined for arbitrary shapes using Minkowski summation. Before the solution is formulated, the FPT problem is discussed.

\section{First-passage time}
The conflict prediction solution presented in this article is inspired by the \emph{first-passage time (FPT) problem} \cite{darling1953}. This problem can be expressed as: given a point mass under the influence of noise -- a stochastic process -- when will this point mass cross or hit a given boundary for the first time? Historically, the problem is well known for its formulation in the context of Brownian diffusion processes \cite{darling1953}. 

The FPT $\tau_{\bm{\mathcal{C}}}$, with reference to a conflict boundary (CB) set $\bm{\mathcal{C}}$ which bounds a conflict region $\bm{\mathcal{D}}$, $\bm{\mathcal{C}} \in \bm{\mathcal{D}}$, in $n$ dimensions, can be expressed as \cite{hittingBessel3Ryznar2013}
\begin{align}
\label{first_P_def}
\tau_{\bm{\mathcal{C}}}(\omega) = \text{inf}\ \{t \in \mathbb{R}^+\  |\ \mathbf{R}(t, \omega) \in \bm{\mathcal{C}} \},
\end{align}
where $\text{inf}$ represents the \emph{infimum function} \cite{hittingBessel3Ryznar2013}. $\mathbf{R}$ is assumed to be a continuous process without discontinuities, therefore the conflict region $\bm{\mathcal{D}}$ cannot be entered without the vehicle touching the CB $\bm{\mathcal{C}}$. The outcome $\omega \in \Omega$ can be thought of as an indicator function for a given stochastically generated trajectory $\mathbf{r}(t)$ \cite{van2009fastMyP28}. The first-passage time is therefore a mapping of the stochastic outcomes to a nonnegative time $\tau_{\bm{\mathcal{C}}} \in \mathbb{R}^+$,  $\tau_{\bm{\mathcal{C}}}: \Omega \to [0,\ \infty) $.
 
 \subsection{First-passage time as a conflict metric}
\vspace{-0.2cm}
The FPT is a useful metric for conflict prediction because it can be used to model the uncertainty of crossing a CB for the first time. The FPT can be used to determine the probability of a conflict event $\mathbb{P} _C$ by integrating the first-passage time distribution (FPTD). This probability, over a prediction interval of $[0,t]$, can be expressed as
 \begin{align}
\label{prob_def}
\mathbb{P} _C(t) =  \bigintssss_{0}^{t} f_{\tau}(s)\ ds,
\end{align}
where $f_{\tau}(s)$ denotes the FPTD and $s$ is an integration variable.
 \subsection{Modelling first-passage time distribution}

Modelling the FPTD for the purpose of conflict prediction can be difficult, given that the distribution is a function of the vehicle motion model, as well as the shape of the CB. As the CB shape becomes more arbitrary and higher dimensional, the FPTD becomes more difficult to describe analytically. In the field of stochastic processes, much effort has been put into 1-D cases \cite{hitting_bessel2}, where the CB is a 1-D point. However, for the purpose of autonomous navigation, a 2-D problem definition (at minimum) is required in order to be useful. FPTDs have been derived for 2-D cases  \cite{iyengar1985hitting10}, but they tend to not have closed-form solutions \cite{iyengar1985hitting11}. A CB consisting of two perpendicular lines has been used to produce an analytical solution \cite{1979Buckholtz}, but only for specific choices of correlation coefficients \cite{HaozheShan2019}. These CBs are also quite restrictive and therefore not as useful in general. Given these challenges, a FPT derivation that yields an analytical solution, while not being restrictive in application, is desirable. The solution formulated in this article uses a FPT derivation in the context of reduced-order dynamics (dimension-reduction), that both has an analytical solution and is modular in application. 
\section{General solution formulation}
\begin{figure}[!htb]
   \centering
   \includegraphics[height=5.6cm]{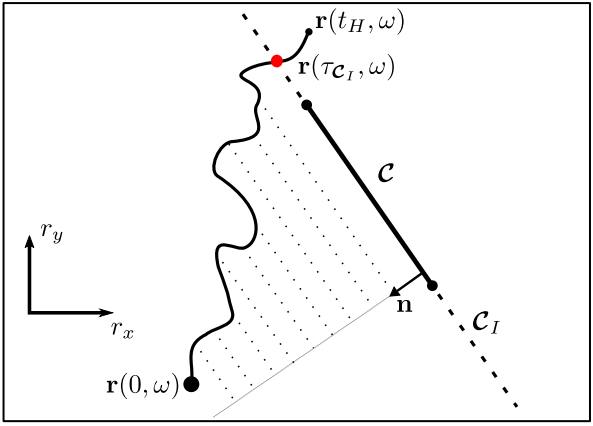}
   \caption[X]{Illustration of dimension-reduction in 2-D, mapping the stochastic process to 1-D. Shown is a sample trajectory, where the red dot is the point of conflict on the trajectory at time $\tau_{\bm{\mathcal{C}}_I}$, and $t_H$ denotes the prediction period (lookahead time). The stochastic process is reduced to a 1-D process that is normal to the line $\bm{\mathcal{C}}_I$, $\mathbf{n}$ being a unit vector normal to and pointing away from $\bm{\mathcal{C}}_I$. This mapping is illustrated with dotted lines. $\bm{\mathcal{C}}_I$ denotes an infinite line, represented by a dashed line, while $\bm{\mathcal{C}}$ denotes the segment of interest, $\bm{\mathcal{C}} \subset \bm{\mathcal{C}}_I$.
  \label{fig5}}
\end{figure}

In this section, we outline our approach and present our solution formulation, which makes use of the FPT. Our solution follows a dimension-reduction approach, illustrated by Fig.~\ref{fig5}, resulting in significant computational savings. Our dimension-reduction approach calculates the probability of conflict for a straight line segment that forms a section of a CB in 2-D. If a given 2-D CB can be approximated with a set of line segments, then the conflict probability can be determined for each line segment. These results can then be summed, yielding an approximation of the probability of conflict.

The stochastic process describing the vehicle's motion can be reduced to a 1-D process:
 \begin{align}
\label{math_work1}
R_\mathbf{n}(t,\omega) = \mathbf{R} (t,\omega) \cdot \mathbf{n},
\end{align}
where $\mathbf{n}$ is defined as a unit vector normal to the CB. The dimension reduction of Equation \ref{math_work1} represents the general case, whereas dimension reduction results for a specific class of stochastic processes are presented in Section 5. The resultant 1-D process is therefore normal to the conflict line, and is treated as a FPT problem. This 1-D conflict problem is equivalent to a 2-D conflict problem relative to an infinite line $\bm{\mathcal{C}}_I$. However, integrating the FPTD $f_{\tau_{\bm{\mathcal{C}}_I}}$ does not yield a useful result, since we are not interested in determining the probability of crossing an infinite line $\bm{\mathcal{C}}_I$, but only a line segment $\bm{\mathcal{C}}$. To solve this problem we use the product of the FPTD and the state uncertainty, and integrate over the subset of interest $\bm{\mathcal{C}}$, as described in the next subsection. 

\subsection{Derivation of probability of conflict using FPT and state uncertainty}
This subsection presents the derivation of the proposed method to calculate the conflict probability, which is used with the FPTD derivation in the next section. The probability of conflict over the prediction period $t_H$, relative to the CB $\bm{\mathcal{C}}_I$, can be expressed as
 \begin{align}
\label{newDer1}
\mathbb{P}^{\bm{\mathcal{C}}_I}_C = \int_{0}^{t_H} f_{\tau_{\bm{\mathcal{C}}_I}}(t) \ dt.
\end{align}
The FPTD $f_{\tau_{\bm{\mathcal{C}}_I}}$ in Equation \ref{newDer1} can be expressed as the marginal of a joint distribution $f_{\mathbf{R}(t),\tau }$ over the vehicle position and FPT as
 \begin{align}
\label{newDer3}
f_{\tau_{\bm{\mathcal{C}}_I}}(t) = \int_{\mathbb{R}^2} f_{\mathbf{R}(t), \tau} \big( \mathbf{r}(t), t \big) \ d \mathbf{r}.
\end{align}
Since $\tau = t$ means that the vehicle is on the CB, $\mathbf{r}(t) \in \bm{\mathcal{C}}_I$, Equation \ref{newDer3} can be expressed as
 \begin{align}
\label{newDer4}
f_{\tau_{\bm{\mathcal{C}}_I}}(t) = \int_{\mathbb{R}^2 \in \bm{\mathcal{C}}_I} f_{\mathbf{R}(t), \tau} \big( \mathbf{r}(t), t \big) \ d \mathbf{r}.
\end{align}
Equation \ref{newDer1} can therefore be expressed as
\vspace{-0.3cm} \allowdisplaybreaks
 \begin{align}
\label{newDer5}
\mathbb{P}^{\bm{\mathcal{C}}_I}_C &=  \int_{0}^{t_H} f_{\tau_{\bm{\mathcal{C}}_I}}(t) \ dt \\
\mathbb{P}^{\bm{\mathcal{C}}_I}_C &= \int_{0}^{t_H}  \int_{\mathbb{R}^2 \in \bm{\mathcal{C}}} f_{\mathbf{R}(t), \tau} \big( \mathbf{r}(t), t \big) \ d \mathbf{r} \ dt \nonumber \\
&+ \int_{0}^{t_H}  \int_{\mathbb{R}^2 \in \ \bm{\mathcal{C}}_{I} \backslash \bm{\mathcal{C}}} f_{\mathbf{R}(t), \tau} \big( \mathbf{r}(t), t \big) \ d \mathbf{r} \ dt.
\end{align} 
The spatial integral is separated over two mutually exclusive sets: the line segment of interest $\bm{\mathcal{C}}$, and the remainder of the infinite line $\bm{\mathcal{C}}_{I} \backslash \bm{\mathcal{C}}$. For the purpose of conflict prediction, we are interested in counting the first crossings of $\bm{\mathcal{C}}_{I}$ over the subset $\bm{\mathcal{C}}$, therefore we set
\begin{align}
\label{newDer6}
\mathbb{P}^{\bm{\mathcal{C}}}_C = \int_{0}^{t_H}  \int_{\mathbb{R}^2 \in \bm{\mathcal{C}}} f_{\mathbf{R}(t), \tau} \big( \mathbf{r}(t), t \big) \ d \mathbf{r} \ dt.
\end{align}
Using the definition of conditional probability distributions, Equation \ref{newDer6} can be expressed as
\begin{align}
\label{newDer7}
\mathbb{P}^{\bm{\mathcal{C}}}_C = \int_{0}^{t_H}  \int_{\mathbb{R}^2 \in \bm{\mathcal{C}}} f_{\mathbf{R}(t)} \big( \mathbf{r}(t)\ | \tau = t \big) f_{\tau_{\bm{\mathcal{C}}_I}}(t) \ d \mathbf{r} \ dt.
\end{align}
The PDF $f_{\mathbf{R}(t)} \big( \mathbf{r}(t)\ | \tau = t \big) $  is called the first-crossing distribution (FCD), denoted by $f_{\Phi}(\mathbf{r}(t))$. The first crossing $\Phi$ is the position on $\bm{\mathcal{C}}_I$ where the vehicle crosses the CB for the first time. We now introduce an important approximation: the FCD is similar to the conditional state distribution, conditioned on the CB. In most cases this approximation is accurate due to the fact that at the FPT, the vehicle is on the CB by definition, assuming absolutely continuous trajectories. However, this substitution is not exact, given that the vehicle could be on the CB after the FPT (second crossings etc.). However, we expect that it is unlikely for the vehicle to return to the CB after the FPT in typical autonomous navigation applications. This approximation can be expressed as
\begin{align}
\label{newDer8}
\nonumber\!\!\!\! f_{\Phi}(\mathbf{r}(t)) &= f_{\mathbf{R}(t)} \big( \mathbf{r}(t) |\ \tau = t \big) \\
&\approx f_{\mathbf{R}(t)} \big( \mathbf{r}(t) |\ \mathbf{r}(t)\in \bm{\mathcal{C}}_I \big)= f_{\Phi}(\mathbf{r}(t))^*,
\end{align}
for short and results in an approximation to the probability of conflict:
\begin{align}
\label{newDer9}
\nonumber \mathbb{P}^{\bm{\mathcal{C}}}_C &\approx \int_{0}^{t_H}  \int_{\mathbb{R}^2 \in \bm{\mathcal{C}}} f_{\mathbf{R}(t)} \big( \mathbf{r}(t)\ | \mathbf{r}(t) \in \bm{\mathcal{C}}_I \big) f_{\tau_{\bm{\mathcal{C}}_I}}(t) \ d \mathbf{r} \ dt\\
&= \int_{0}^{t_H} f_{\tau_{\bm{\mathcal{C}}_I}}(t) \int_{\mathbb{R}^2 \in \bm{\mathcal{C}}} f_{\Phi}(\mathbf{r}(t))^*  \ d \mathbf{r} \ dt.
\end{align}
At each moment $t$ in time, the conditional state distribution is conditioned on and integrated over the CB. This result is then weighed by the likelihood that $t$ is in fact the FPT. Therefore, the FPTD acts like a time-dependent weighting function. Once the inner integral in Equation \ref{newDer9} is calculated, the resultant expression consists of a single numerical integral over time. Equation \ref{newDer9} uses a 1-D FPTD which is derived in Section 6. The inner integral in Equation \ref{newDer9} can be calculated analytically using the conditional state cumulative distribution function (CDF) for a specific class of models, outlined in the next section.

\section{State space models}
We are interested in stochastic processes that will result in a closed-form expression for the inner integral in Equation \ref{newDer9}, as well as an analytical result for the FPTD in Section 6. Useful models that satisfy these requirements are a class of linear, Gaussian, time-invariant, continuous models. We focus on these models for the rest of the article and refer to this class of models as the vehicle motion model. These models are often used to approximate vehicle motion \cite{NewAtricle1,jones2003realMyP10,NewAtricle2}. Such a model can be expressed in terms of a stochastic differential equation \cite{grimble1988optimal}
\vspace{-0.2cm} 
\begin{align}
\label{OL_system}
\!\!\! \mathbf{\dot{X}}(t, \omega)\! =\! \mathbf{A}(t) \mathbf{X}(t, \omega) + \mathbf{B}(t) \mathbf{u}(t) + \mathbf{B}_{\eta}(t) \pmb{\eta}(t, \omega),
\end{align}
where $\mathbf{X}(t,\ \omega)$ is a vector of stochastic processes.  $\pmb{\eta}(t,\ \omega)$ is a vector of zero-mean, Gaussian, white noise. $\mathbf{A}(t)$ denotes the state matrix, $\mathbf{u}(t)$ the control input, $\mathbf{B}(t)$ the input matrix, and $\mathbf{B}_{\eta}(t)$ the noise input matrix. The solution to Equation \ref{OL_system}, starting from $t=0$, is \cite{grimble1988optimal,borrie1992stochastic} 
\vspace{-0.3cm} 
\begin{align}
\label{OL_system_solve}
\mathbf{X}(t,\ \omega) &= \pmb{\Psi}(t)\ \mathbf{X}(0,\ \omega) + \bigintssss_{0}^{t_H} \pmb{\Psi}(t)\ \mathbf{B}(t)\ \mathbf{u}(t)\ dt \nonumber \\ 
&+ \bigintssss_{0}^{t_H} \mathbf{B}_{\eta}(t)\ d \textbf{W}(t,\ \omega),
\end{align}
where $\pmb{\Psi}(t)$ denotes the state transition matrix. The final integral $\bigintssss_{ }^{ } \cdot  \ d \textbf{W}$ denotes an Ito-integral over a differential increment in an extended Wiener process \cite{borrie1992stochastic}, with a diffusion matrix $\textbf{Q}(t)$. The model consists of linear transforms and Gaussian distributed noise; if the initial states are Gaussian distributed, $\mathbf{X}(0,\ \omega) \sim \mathcal{N} \Big(\ \mathbf{m}(0)\ ,\ \textbf{C} (0)  \Big) $, then all subsequent states are jointly Gaussian distributed \cite{van2010conflictMyP27}, 
 \vspace{-0.2cm}
\begin{align}
\label{NOISE_m}
\mathbf{X}(t,\ \omega) \sim \mathcal{N} \Big(\ \mathbf{m}(t)\ ,\ \textbf{C} (t)  \Big).
\end{align}
The position $\mathbf{R}(t,\ \omega) = [R_X(t,\ \omega),\ R_Y(t,\ \omega)]^T$ and velocity states $\mathbf{V}(t,\ \omega)$ constitute the state vector:
\begin{align}
\label{OL_statesss}
\mathbf{X}(t,\ \omega) = \Big[ \mathbf{R}(t,\ \omega) \ ,\ \mathbf{V}(t,\ \omega) \Big]^T,
\end{align}  
and therefore
\begin{align}
\label{NOISE_m_big}
\! \! \! \!     \mathbf{X}(t,\ \omega) \sim \mathcal{N} \bigg(\ 
\! \! \begin{bmatrix}
\mathbf{m}_{\mathbf{R}}(t,\omega) \\
\mathbf{m}_{\mathbf{V}}(t,\omega)  
\end{bmatrix},\!
 \begin{bmatrix}
 \textbf{C}_{\textbf{R}} (t) & \textbf{C}_{\textbf{RV}} (t) \\
 \textbf{C}^T_{\textbf{RV}} (t) & \textbf{C}_{\textbf{V}} (t)
  \end{bmatrix} \bigg).
\end{align}
Using the model in Equation \ref{OL_system_solve}, the state mean $\mathbf{m}(t)$ and covariance $\textbf{C} (t)$ can be derived analytically. The mean and covariance can also be calculated in a tractable manner if Equation \ref{OL_system} is discretised in terms of $t$. The discrete equivalent model can be used to recursively calculate the mean and covariance at sample times \cite{grimble1988optimal}. 

For a horizontal conflict boundary $\bm{\mathcal{C}}$ with end points $(x_{\bm{\mathcal{C}}1,}y_{\bm{\mathcal{C}}})$ and $(x_{\bm{\mathcal{C}}2},y_{\bm{\mathcal{C}}})$, the conditional state distribution in Equation \ref{newDer9} can be expressed as a Gaussian distribution:
\vspace{-0.3cm} 
\begin{align}
\label{cond_distribution}
f_{\Phi}&(\mathbf{r}(t))^* \sim \mathcal{N} \Big(\ m_{R_X|R_Y}(t) , C_{R_X|R_Y}(t)   \Big)\ \nonumber\\
&=\mathcal{N} \Big(\ m_{R_X}(t)+(y_{\bm{\mathcal{C}}}-m_{R_Y}(t))C_{R_Y R_X}(t)/C_{R_Y}(t) , \nonumber\\ 
&\ \ \ \ \ \ \ \ \  \ \ \ C_{R_X}(t)-C^{2}_{R_Y R_X}(t)/C_{R_Y}(t)  \Big).
\end{align}
 Any problem definition can be transformed so that that the conflict boundary is horizontal, allowing for convenient conditional state mean and variance calculations. Therefore, assuming a horizontal conflict boundary, and using the Gaussian nature of the conditional state distribution, the inner integral in Equation \ref{newDer9} can be calculated using error functions as
\begin{align}
\label{error_inner_integral}
\int_{\mathbb{R}^2 \in \bm{\mathcal{C}}} f_{\Phi}(\mathbf{r}(t))^*  \ d \mathbf{r} &= \frac{1}{2} \bigg[ \text{erf}\Big( \frac{x_{\bm{\mathcal{C}}2}-m_{R_X|R_Y}(t)}{\sqrt{2C_{R_X|R_Y}(t)}} \Big) \nonumber \\ 
&- \text{erf} \Big( \frac{x_{\bm{\mathcal{C}}1}-m_{R_X|R_Y}(t)}{\sqrt{2C_{R_X|R_Y}(t)}} \Big) \bigg].
\end{align}\vspace{-0.5cm}

Using dimension reduction (Equation \ref{math_work1}), and the dot product formulation of the affine transformation of a normal random variable, the resultant Gaussian process (Equation \ref{NOISE_m}) can be shown to reduce to
\begin{align}
\label{NOISE_m_reduced}
R_\mathbf{n} (t,\ \omega) \sim \mathcal{N} \Big(\ \mathbf{m}_{\mathbf{R}}(t) \cdot \mathbf{n}\ ,\   \mathbf{n}^T \textbf{C}_{\textbf{R}} (t) \mathbf{n} \Big), \text{ and}
\end{align}\vspace{-0.7cm}
\begin{align}
\label{NOISE_m_reduced_v}
V_\mathbf{n} (t,\ \omega) \sim \mathcal{N} \Big(\ \mathbf{m}_{\mathbf{V}}(t) \cdot \mathbf{n}\ ,\   \mathbf{n}^T \textbf{C}_{\textbf{V}} (t) \mathbf{n} \Big).
\end{align}\vspace{-0.9cm}

In this way the vehicle's state uncertainty can be modelled in 1-D and used for the FPTD. The derivation of the FPTD in the next section uses the motion model reduced to 1-D (Equations \ref{NOISE_m_reduced} and \ref{NOISE_m_reduced_v}), as described by Equation \ref{math_work1}.
\section{FPTD for state space models}
This section derives the FPTD for linear, Gaussian, 1-D motion models used to characterise $f_{\tau_{\bm{\mathcal{C}}_I}}(t)$ in Equation \ref{newDer9}. Additional assumptions are made that the mean of the velocity is constant, and that the mean position of the vehicle relative to the conflict boundary is a linear function of time. The constant mean velocity makes the model applicable for vehicles that are not accelerating in the mean; instances of acceleration induced by noise do not invalidate this assumption. The mean position being a linear function of time makes the model applicable for any mean motion that can be modelled or approximated as piece-wise straight line segments. The 1-D stochastic process was obtained by reducing the full state stochastic process (Equation \ref{NOISE_m_big}) using dimension-reduction (Equations \ref{NOISE_m_reduced} and \ref{NOISE_m_reduced_v}). The resultant linear dynamics, with Gaussian distributed noise and additional assumptions, can be expressed as a Gaussian process with time-dependent mean and variance as \vspace{-0.2cm} 
\begin{align}
\label{mirrorGaussian122}
R_\mathbf{n}(t,\omega)  &\sim \mathcal{N} \Big(m(t)\ ,\ c(t) \Big) \nonumber \\
&= \mathcal{N} \Big(\mu t + r_\mathbf{n}(0) \ ,\ c(t) \Big) 
= f_{R_{\mathbf{n}}}(r_{\mathbf{n}}(t)).
\end{align}
The mean $m(t)$ is also assumed to be a linear function of time, where $\mu$ denotes the drift factor, which corresponds to the mean of the velocity. The variance is left generic, since its form depends on whether the model uses feedback control (closed-loop), or not (open-loop)

\begin{figure}[!htb]
   \centering
   \includegraphics[height=3.4cm]{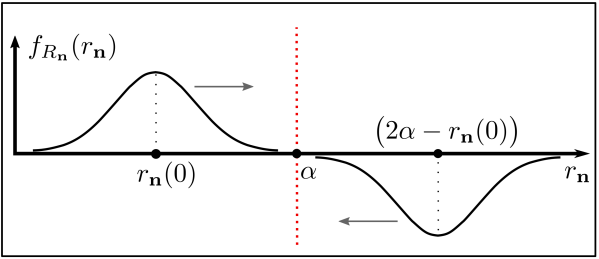}   
   \caption[X]{Illustration of the imaging method \cite{HaozheShan2019}, which satisfies the boundary condition. $\alpha$ denotes the CB and $r_{\mathbf{n}}(0)$ denotes the initial condition. Both means of the Gaussian distributions start at a distance $|\alpha - r_\mathbf{n}(0)|$ from the CB $\alpha$.
  \label{fig6}}
\end{figure}

The FPTD derivation starts by creating an unnormalised distribution $f_{R_\mathbf{n}}^*(r_\mathbf{n} (t)\ | \ \alpha, \ r_\mathbf{n}(0))$ that matches the 1-D position-state distribution $f_{R_\mathbf{n}}$, but also satisfies an absorbing boundary condition: $f_{R_\mathbf{n}}^*(r_\mathbf{n} = \alpha) = 0$.  $\alpha$ denotes the CB point in 1-D. $f_{R_\mathbf{n}}^*$ models the distribution of position states that have not experienced conflict, in contrast to $f_{R_\mathbf{n}}$, which does not account for prior conflict. The distribution $f_{R_\mathbf{n}}^*$ is zero at the CB $\alpha$, also referred to as the \emph{absorbing level} in stochastic process literature \cite{Cox1965}. This distribution will be used to calculate the survival CDF $F_{S}$, which will be used to calculate the FPTD. Fig.~\ref{fig6}. illustrates the approach, which is known as the \emph{method of imaging} \cite{HaozheShan2019,Cox1965}. A negative distribution, which is also spaced a distance of $|\alpha - r_\mathbf{n}(0)|$ from the absorbing point $\alpha$, is used to ensure that the absorbing condition $f_{R_\mathbf{n}}^*(\alpha) = 0$ is met. As the positive distribution state distribution drifts toward the absorbing point from the left, the negative distribution drifts at the same mean rate $|\mu|$ from the right, causing $f_{R_\mathbf{n}}^*(\alpha) = 0\text{,}\ \forall\ t$. 

The resultant distribution can be expressed as \cite{Cox1965} \vspace{-0.5cm}
\begin{align}
\label{mirrorGaussian1}
&f_{R_{\mathbf{n}} }^*(r_{\mathbf{n}} | \ \alpha, \ r_{\mathbf{n}}(0)) = \frac{1}{\sqrt{2\pi c(t)}} \Bigg[\text{exp}\Bigg(\frac{-\big(r_{\mathbf{n}} -m(t)\big)^2}{2 c(t)}\Bigg) \nonumber \\ 
& \ \ \ \ \  \ \ \ - \text{exp}\Bigg(\frac{-\big(r_{\mathbf{n}} -(2\alpha -m(t))\big)^2}{2 c(t)}\Bigg) \Bigg]\text{,}\ r_{\mathbf{n}} \leq \alpha,\\
&\text{or} \ \ f_{R_{\mathbf{n}} }^*(r_{\mathbf{n}} | \ \alpha, \ r_{\mathbf{n}}(0))= 0, \      r_{\mathbf{n}} > \alpha. \nonumber
\end{align}
Note that the variances are the same, but the means are defined in such as way as to cause the mirrored distributions to move toward each other. As they move, their means maintain the same distance from the CB, on both sides of $\alpha$. The distribution $f_{R_{\mathbf{n}} }^*$ is approximately normalised if the Gaussian images are far away from the absorbing point or have a small variance. As the Gaussian images move toward the absorbing point, the distribution $f_{R_{\mathbf{n}} }^*$ becomes less normalised. The rate at which it deviates from being normalised is a function of the mean and variance of of the vehicle. Therefore, Equation \ref{mirrorGaussian1} is a poor approximation for problem definitions with large uncertainty, with means close to the absorbing point.
\vspace{-0.3cm}
\subsection{Survival cumulative distribution}
\vspace{-0.2cm}
The next step is to relate the distribution to the survival metric, survival being defined as not yet having reached the CB $\alpha$. The survival CDF $F_S(t)$, which is related to the FPTD, can be derived as  \cite{Cox1965} \vspace{-0.3cm}
\begin{align}
\label{Surv}
&F_S(t) = \mathbb{P}[R_{\mathbf{n}}<\alpha] = \int_{-\infty}^{\alpha} f_{R_{\mathbf{n}}}^*(r_{\mathbf{n}} | \ \alpha,\ r_{\mathbf{n}} (0)) \ dr_{\mathbf{n}} \\
&= \frac{1}{\sqrt{2\pi c(t)}} \Bigg[ \bigintssss_{-\infty}^{\alpha} \text{exp}\Bigg(\frac{-\big(r_{\mathbf{n}} -m(t)\big)^2}{2 c(t)}\Bigg) dr_{\mathbf{n}} \nonumber \\ & \ \ \ \ \ \ - \bigintssss_{-\infty}^{\alpha} \text{exp}\Bigg(\frac{-\big(r_{\mathbf{n}} -(2\alpha -m(t))\big)^2}{2 c(t)}\Bigg) dr_{\mathbf{n}} \Bigg]. \label{SurV}
\end{align}
The probability mass of $f_{R_{\mathbf{n}}}^*$ corresponds to the portion of $f_{R_{\mathbf{n}}}$ that has not yet experienced conflict prior to the current moment $t$, and therefore represents the survival probability. Using variable substitutions $u = (r_{\mathbf{n}}-m(t))/ \sqrt{2c(t)}$ and $v = (r_{\mathbf{n}}-2\alpha +m(t))/ \sqrt{2c(t)}$, $c(t) \neq 0, \ \forall \ t$, and dividing both of the integration bounds at zero, both integrals over $[- \infty,0]$ cancel out, and Equation \ref{SurV} reduces to
\begin{align}
F_S(&t) =  \frac{1}{\sqrt{\pi}} \Bigg[ \int_{0}^{u(\alpha)} e^{-u^2} du  - \int_{0}^{v(\alpha)} e^{-v^2} dv \Bigg].
\label{Surv2222}
\end{align}
Equation \ref{Surv2222} can now be expressed as
\begin{align}
\label{Surv2}
\!\!F_S&(t)\!\! =\!\! \frac{1}{2} \Bigg[ \text{erf}\bigg(\frac{\alpha-m(t)}{\sqrt{2c(t)}} \bigg) - \text{erf}\bigg(\frac{-(\alpha -m(t))}{\sqrt{2c(t)}}\bigg)  \Bigg].
\end{align}
Given that the error function is an uneven function, $\text{erf}(-x) = -\text{erf}(x)$, the survival CDF reduces to
\begin{equation}
\label{O-U-Gaus_ss2222}   
F_S(t) = \text{erf}\bigg(\frac{\alpha-m(t)}{\sqrt{2c(t)}} \bigg)\text{,}\ \alpha - m(t) > 0.
\end{equation} 
The survival CDF can be related to the FPTD $f_{\tau}(t)$ as \cite{Cox1965} 
\begin{equation}
\label{surv_3_FHT}
f_{\tau}(\ t) = - \frac{dF_S(t)}{dt} .
\end{equation}
Using the expression for the survival CDF, Equation \ref{O-U-Gaus_ss2222}, the FPTD $f_{\tau}(\ t)$ can be derived using the chain rule as
\begin{align}
\label{FHT_appDer}
f_{\tau}(t) &= - \frac{dF_S(t)}{dt} \nonumber =  \sqrt{\frac{2}{c(t)\pi}} \text{exp} \Bigg( \frac{-\Big(\alpha - m(t)\Big)^2}{2c(t)}   \Bigg) \times \nonumber \\ & \ \ \ \ \ \ \ \ \ \ \ \ \ \ \ \ \ \ \  \ \ \ \ \ \ \ \ \  \Bigg( \frac{(\alpha - m(t))\dot{c}(t)}{2c(t)} + \mu\Bigg),
\end{align}
noting that $\dot{m}(t) = \mu$; the mean changes linearly as defined by Equation \ref{mirrorGaussian122}. The limitation of the method can be seen in Equation \ref{FHT_appDer}: the FPTD collapses if both $\mu$ and $\dot{c}$ are zero, or is invalid if the second factor is negative. After more specific models are explored next, the negative factor is only an issue if the vehicle is not moving toward the conflict boundary. A conflict prediction algorithm could inspect the mean movement (relative to the CB) to determine if the method is useful, bearing in mind that moving away from the CB poses minimal risk in most encounter scenarios.
\vspace{-0.15cm}
\subsection{Closed-loop scenario}
\vspace{-0.15cm}
The result in Equation \ref{FHT_appDer} can be further refined by choosing appropriate expressions for the variance $c(t)$. For a closed-loop system using a time-invariant controller, the variance is regulated by means of feedback control, causing it to settle at a steady-state value $c_{ss}$ after a while. From this point the derivative is zero $\dot{c}(t) = 0$. For typical applications the variance would settle quickly; taking advantage of a constant (zero) variance reduces Equation \ref{FHT_appDer} to \vspace{-0.2cm}
\begin{equation}
\label{surv_3_final_almost}
f_{\tau}( t) = \frac{2\mu}{\sqrt{2\pi c_{ss}}} \text{exp} \Bigg( \frac{-\Big(\alpha - m(t)\Big)^2}{2c_{ss}}   \Bigg),
\end{equation}
for closed-loop systems. Integrating $f_{\tau}( t)$ over all time $0 \rightarrow \infty$ results in \vspace{-0.5cm}
\begin{equation}
\label{surv_3_int}
\int_{0}^{\infty} f_{\tau}( t) dt = 2,
\end{equation}
due to the choice of $f_{R_{\mathbf{n}}}^*$ in Equation \ref{mirrorGaussian1}. The FPTD is normalised, resulting in  \vspace{-0.5cm}
\begin{equation}
\label{surv_3_final_positive}
f_{\tau}( t) = \frac{\mu}{\sqrt{2\pi c_{ss}}} \text{exp} \Bigg( \frac{-\Big(\alpha - m(t)\Big)^2}{2c_{ss}}   \Bigg) .
\end{equation}
Following the same derivation, but changing the illustration in Fig.~\ref{fig6} to a $\alpha<r_{\mathbf{n}}(0)$ and $\mu < 0$ scenario produces
\vspace{-0.6cm} 
\begin{equation}
\label{surv_3_final_neg}
f_{\tau}( t) = \frac{-\mu}{\sqrt{2\pi c_{ss}}} \text{exp} \Bigg( \frac{-\Big(\alpha - m(t)\Big)^2}{2c_{ss}}   \Bigg)
\end{equation}
after normalisation. These two results can be put together as \vspace{-0.5cm}
\begin{equation}
\label{surv_3_final_final}
f_{\tau}( t)= \frac{|\mu|}{\sqrt{2\pi c_{ss}}} \text{exp} \Bigg( \frac{-\Big(\alpha - m(t)\Big)^2}{2c_{ss}}   \Bigg),
\end{equation}
which is the final form of the FPTD of closed-loop dynamics, with a linear mean as defined by Equation \ref{mirrorGaussian122}.
\vspace{-0.15cm}
\subsection{Open-loop scenario}
\vspace{-0.15cm}
An open-loop model can be used to model dynamics that do not have any feedback corrections. For example, Yang et al. \cite{yang2004realMyP31} used an open-loop model to represent the timing uncertainty of a flight manoeuvre used by a pilot changing their flight path. For an open-loop system, the state distribution can be defined as \vspace{-0.2cm}
\begin{align}
\label{OL-ss-model}
R_{\mathbf{n}}(t,\omega) &\sim \mathcal{N} \Big(m(t)\ ,\ c(t) \Big) \nonumber \\
&= \mathcal{N} \Big(\mu t + r_{\mathbf{n}}(0) \ ,\ \frac{1}{3}\sigma^2 t^3 \Big),
\end{align}
where the variance is the closed-form solution of a linear state space model with zero-mean Gaussian noise, 
with a constant mean velocity and an initial zero covariance. $\sigma$ denotes the white noise strength. The expression given in Equation 
\ref{OL-ss-model} is a useful starting point; however, the work that follows is without loss of generality, since the third order time $t^3$ is the governing factor in the derivation. The FPTD can be determined from the result in Equation \ref{FHT_appDer}: \vspace{-0.2cm}
\begin{align}
\label{FHTD_OL_cov3}
&f_{\tau}(t) = - \frac{dF_S(t)}{dt} \nonumber \\ 
&= \frac{-2}{\sqrt{\pi}} \text{exp} \Bigg( \frac{-\big(\alpha - m(t)\big)^2}{2c(t)}   \Bigg) \frac{d}{dt}\Bigg( \frac{(\alpha - m(t))}{\sqrt{(2c(t))}}\Bigg).
\end{align}
Using the model as defined by Equation \ref{OL-ss-model} results in
\begin{align}
\label{FHTD_OL_cov3_2}
f_{\tau}( t) &= \text{exp} \Bigg( \frac{- \big( \alpha - m(t) \big)^2}{2c(t)}   \Bigg) \frac{\sqrt{3 t} (3a - \mu t)}{\sqrt{2 \pi} \sigma t^3},
\end{align}
where $a = \alpha - r_{\mathbf{n}}(0)$. The distribution is only valid for $t < \frac{3a}{\mu} $; outside this range the CDF is no longer monotonically increasing. Equation \ref{FHTD_OL_cov3_2} needs to be slightly altered in two ways; First, Equation \ref{FHTD_OL_cov3_2} needs to be altered to work for encounter scenarios where $a$ and $\mu$ are negative $a < 0$, $\mu < 0$. This alteration can be done with an absolute value function. Second, the expression needs a normalisation factor so that the CDF tends to unity. This factor is the same as the closed-loop case. Equation \ref{FHTD_OL_cov3} can be expressed in its final form as
\begin{align}
\label{FHTD_OL_cov3_final}
f_{\tau}( t) &= \text{exp} \Bigg( \frac{- \big( \alpha - m(t) \big)^2}{2c(t)}   \Bigg) \times   \frac{\sqrt{3 t} \ |3a - \mu t|}{2 \sqrt{2 \pi} \sigma t^3},
\end{align}
for $t < \frac{3 a}{\mu}$.Having derived the FPTD for open- and closed-loop motion models as defined by Equation \ref{mirrorGaussian122}, Equation \ref{newDer9} can be used to approximate the probability of conflict. 
\vspace{-0.3cm}
\section{Simulation experiments} 
\vspace{-0.1cm}
This section contains simulations of open-loop and closed-loop scenarios illustrating the advantage of the proposed method. The proposed method is compared to three existing methods: a probability flow (PF) method by van Daalen and Jones \cite{van2009fastMyP28}, a more recent PF variation by Park and Kim \cite{flow_As_2017MyP22}, and an instantaneous conflict probability (ICP) method by Pour et al. \cite{pour2019probability}. These methods were chosen for their applicability to the testing scenarios and their execution speed. 

The PF method of Park and Kim was slightly altered due to their Wiener process motion model (Equations 1 and 2 in \cite{flow_As_2017MyP22}). The drift contribution to PF is not defined so as to constrain its contribution to only drift directed toward the CB, which consequently allows for negative drift PF contributions. Negative PF is ill defined in a conflict predication framework since a vehicle cannot recover after experiencing conflict. The published method uses the constant velocity (Wiener drift) normal to the CB $(\mathbf{v} \cdot \mathbf{n})$ to calculate the drift contribution (Equation 10 of \cite{flow_As_2017MyP22}), regardless of direction. Given these discrepancies, the velocity normal to the CB $(\mathbf{v} \cdot \mathbf{n})$ in Equation 10 of \cite{flow_As_2017MyP22} is replaced by the unconditional velocity mean normal to the CB ($\beta$), calculated using velocities oriented toward the CB: 
\vspace{-0.2cm} 
\begin{align}
\label{Park_1}
\beta = \int_{- \infty}^{0} f_{V_\mathbf{n}}(v_\mathbf{n}) v_{\mathbf{n}} \ dv_\mathbf{n}.
\end{align}
Negative velocities are considered since positive velocities denotes PF away from the CB. A second method was also implemented that corresponds to the publication version; the velocity mean was used as the publication Wiener drift counterpart, since the velocity mean of a Wiener process is equal to its drift. Both the implementation results of Park and Kim's published version and the altered version are presented. However, both are altered to ensure that the drift contribution's velocity points toward the CB; in the publication version the normal velocity is limited using an if statement, while the altered version uses Equation \ref{Park_1}. 

In a recent paper, Pour et al. proposed a method to calculate the ICP by approximating the conflict region with rectangular elements, after transforming the state space to achieve an uncoupled joint distribution \cite{pour2019probability}. 
In a follow-up paper the authors adapted their conflict region approximation by using a spatial multi-resolution scheme \cite{seyedipour2021efficient}. The new adaptation dynamically changes the conflict region approximation as a function of the relative position uncertainty. This article focusses on the first paper (original method) in order to control and manually vary the region approximation accuracy. 

ICP is not synonymous with conflict probability and requires further processing to produce a conflict probability result. Three conflict probability results are derived from the published ICP algorithm: the maximum ICP (lower-bound estimate) \cite{Erzberger_paielli_1997MyP21}, and two forms of ICP accumulation \cite{jones2006ArticleMyProp11}. The first assumes no prior conflict before the last discrete accumulation, and can be expressed using discrete notation, with $\mathbb{P}_C[0] =0$ and $k \geq 0$, as 
\begin{align}
\label{Pour_1}
\mathbb{P}_C[k+1] =  \mathbb{P}_C[k] + \text{ICP}[k+1] (1-\mathbb{P}_C[k]).
\end{align}
The second includes all ICP from the beginning, with $\mathbb{P}_C[0] =0$ and $k \geq 0$, expressed as
\vspace{-0.2cm} 
\begin{align}
\label{Pour_2}
\mathbb{P}_C[k+1] =  \mathbb{P}_C[k] + \text{ICP}[k+1]\prod_{i=0}^{k} (1-\mathbb{P}_C[i]).
\end{align}
Note the method of Pour et al. is only used for the first simulation since the method is only defined for circular or elliptical conflict regions. 

The actual conflict probability for each scenario is calculated using Monte Carlo simulations, which are used to evaluate each method's accuracy. For each simulation, normal vectors on the CB and pointing away from the conflict region are used to determine whether the vehicle is moving toward the CB.
All tests were executed on a Linux Mint 64-bit operating system, with 16 GB RAM and an Intel Core i5-4590 CPU @ 3.30 GHz$\times$4. Conflict probability algorithms were implemented in C and called in Python by means of wrappers. 
\vspace{-0.1cm}
\subsection{Open-loop experiment} 
\begin{figure}[!htb]
  \centering               
   \includegraphics[width=8.4cm]{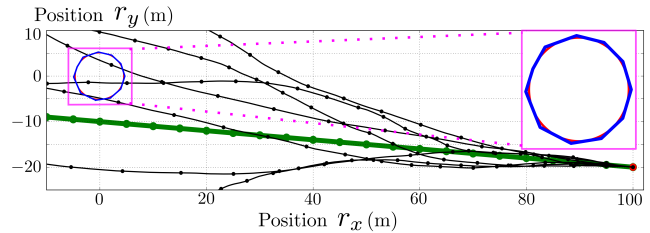}   \vspace{-0.4cm}   
   \caption[XX]{2-D open-loop simulation. The CB (red) is a circle at the origin with radius of 5. Six straight line segments (blue) can used to represent the CB on the right side. The nominal trajectory (green) corresponds to the mean path of a Gaussian process. Sample trajectories (black) are also displayed.
  \label{fig7}} 
\end{figure}

\vspace{-0.2cm}
In the experiment for the open-loop scenario, the vehicle starts at $\mathbf{r} = (100\ \text{m},-20\ \text{m})$, as depicted by Fig.~\ref{fig7}, with an initial covariance of zero. The vehicle has a constant mean velocity of $(-10\ \text{m/s},1\ \text{m/s})$, and its acceleration is stimulated by noise with a diffusion matrix of $\textbf{Q} = \text{diag}(2.2^2,1.58^2)$. The total prediction period $t_H$ is $15$ seconds, with a sampling period of $15$ ms. 
\vspace{-0.2cm} 

PF involves numerical integration along the CB in state space, therefore a partition interval for the circle circumference must be chosen. Since PF involves spatial numerical integration, and the proposed method does not, two results are presented for PF for comparison. In the first case, the circle is partitioned into 20 intervals, while the second is partitioned into 15 intervals. These two partitions are used for all PF methods. The ICP method of Pour et al. involves partitioning the conflict region using rectangular elements, likewise two results are presented for comparison; the region is partitioned using 15 elements for the first case and 20 elements for the second case. Note that partitioning the CB is distinct from partitioning the conflict region into rectangular elements. For all methods, smaller partition intervals are likely to yield more accurate answer at the cost of additional computation. Table \ref{myTab_OpenLoop} summarises the simulation results. 

\begin{center}
\begin{table}[ht]
    \centering
     \caption{Collation of open-loop simulation results. The partition sizes are shown in brackets; the corresponding results coincide with the partition size order. The published version of Park and Kim's method is indicated with ``(P)", and the altered version is indicated with ``(A)". ``Accumulation" has been shortened to ``Acc.". }
    \label{myTab_OpenLoop}
    \begin{tabular}{|m{3.1cm} || m{1.95cm} | m{1.95cm} |}
    \Xhline{3\arrayrulewidth} 
      \centering \textbf{Methods}  &  \centering Average\footnotemark[4] run time (ms)  &   \centering Probability of conflict (\%)  \cr \Xhline{3\arrayrulewidth} 
      \centering Monte Carlo\footnotemark[3] & \centering N/A & \centering  11.344  \cr    \hline
      \centering Proposed method &  \centering 1.212 &  \centering 11.359 \cr    \hline   
       \centering van Daalen and Jones \cite{van2009fastMyP28} (20/15) &   \centering 2.675/2.022  &  \centering 11.402/11.396    \cr  \hline 
     \centering Park and Kim (P) \cite{flow_As_2017MyP22} (20/15) &   \centering  2.493/1.909  &  \centering 9.939/9.931    \cr \hline 
     \centering Park and Kim (A) \cite{flow_As_2017MyP22} (20/15) &   \centering  3.540/2.714  &  \centering 11.480/11.480    \cr \hline 
     \centering Max. ICP, Pour et al. \cite{pour2019probability} (20/15) &   \centering  0.284/0.217  &  \centering 1.375/1.366    \cr \hline 
     \centering ICP Acc. Eq. \ref{Pour_1}, Pour et al. \cite{pour2019probability} (20/15) &   \centering  0.284/0.217  &  \centering 37.927/37.735    \cr \hline 
     \centering ICP Acc. Eq. \ref{Pour_2}, Pour et al. \cite{pour2019probability} (20/15) &   \centering  0.284/0.217  &  \centering 14.743/14.697    \cr
       \Xhline{3\arrayrulewidth} 
    \end{tabular}
   
\end{table}
\end{center} \footnotetext[3]{Simulation performed with 4,414,427 sample trajectories.} \footnotetext[4]{Average calculated with 10,000 simulation runs.} 
\vspace{-0.5cm}
 
The execution time of the proposed method is faster than most PF results by a factor of 2. This is largely due to the computational cost of numerical integration; the proposed method requires the calculation of a single, temporal numerical integral, while PF requires both temporal and spatial numerical integrals. The ICP calculations were faster than the proposed method, but at the cost of accuracy. The ICP accumulation results were obtained by calculating and accumulating the ICP every 150 ms. The lower accumulation rate is used to ensure that the accumulated result does not drastically compound and saturate over time. The ICP accumulation tends to saturate over time due to the assumption that the vehicle state distribution is unchanged by previous conflict; the impact of this assumption becomes more pronounced if the accumulation points are within close proximity in time. These problems can be regarded as a short-coming of ICP methods: they do not directly translate to a calculation of total vehicle conflict probability without limiting assumptions, like the maximum ICP, or further data processing decisions such as a lower accumulation rate. The absolute error of all the PF simulation results are relatively small, with the largest absolute error being 1.413\% and smallest absolute error being 0.052\%. The proposed method resulted in an absolute error of 0.015\%.

\subsection{Closed-loop experiment} 
\begin{figure}[!htb]
 \centering              
  \includegraphics[width=8.4cm]{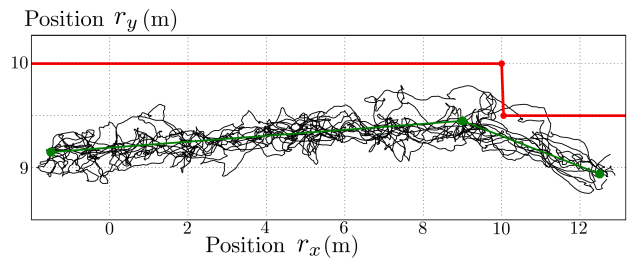}   \vspace{-0.4cm}      
   \caption[X]{2-D closed-loop simulation. The CB (red) consists of three line segments. The nominal trajectory (green) corresponds to the mean path of a Gaussian process. Sample trajectories (black) are also displayed.
  \label{fig8}}
\end{figure}

\vspace{-0.2cm} 
In the closed-loop experiment scenario, the vehicle starts at $\mathbf{r} = (-1.5\ \text{m}, 9.15\ \text{m})$, as depicted by Fig.~\ref{fig8}, with an initial covariance of zero. The vehicle moves in two stages; first, with a constant mean velocity magnitude of $1$ m/s to position $(9\ \text{m}, 9.45\ \text{m})$. Second, from $(9\ \text{m}, 9.45\ \text{m})$ the vehicle moves with a constant mean velocity magnitude of $1.6$ m/s to position $(12.5\ \text{m}, 8.94\ \text{m})$. Two straight-line manoeuvres are followed, with feedback control tracking the planned path. The total prediction period is $13.73$ seconds, with a sampling period of $10$ ms. During both manoeuvres the acceleration is stimulated by noise with a diffusion matrix of $\textbf{Q} = \text{diag}(7.5^2,2.4^2)$.  The proposed method calculates the steady-state covariance using the feedback gains and noise variances. Therefore, an approximation is introduced that there are no transient dynamics. The simulation results are shown in Table \ref{myTab_ClosedLoop}. \vspace{0.05cm}

\vspace{-0.3cm} 

Three results are shown for each PF simulation; for the first result, each line is partitioned into intervals of 0.05 m, for the second result each line is partitioned into intervals of 0.1 m, and for the third result each line is partitioned into intervals of 0.15 m. The proposed method makes an absolute error of 1.3\%. The largest PF result absolute error is 9.206\% and the smallest PF result absolute error is 0.149\%. The most substantial advantage of the proposed method is the computational savings: All PF results are at least an order of magnitude more expensive. This is due to the large amount of CB partitioning needed, inflating the spatial integral computational cost. The total spatial integral cost accumulates drastically since the calculation is performed for each time step.
\begin{center}
\begin{table}[ht]
    \centering
    \caption{Collation of closed-loop simulation results. Partition interval lengths (m) are shown in brackets; the corresponding results coincide with the interval length order. The published version of Park and Kim's method is indicated with ``(P)", and the altered version is indicated with ``(A).".}
    \label{myTab_ClosedLoop}
    \begin{tabular}{|m{3cm} || m{2cm} | m{2cm} |}
    \Xhline{3\arrayrulewidth} 
      \centering \textbf{Methods}  &  \centering Average\footnotemark[4] run time (ms)  &   \centering Probability of conflict (\%)  \cr \Xhline{3\arrayrulewidth} 
      \centering Monte Carlo\footnotemark[5] & \centering N/A & \centering  10.231 \cr    \hline
      \centering Proposed method &  \centering 0.278 &  \centering 8.931 \cr    \hline   
       \centering van Daalen and Jones \cite{van2009fastMyP28} (0.05/0.1/0.15) &   \centering 107.561/ 26.491/ 17.640  &  \centering 12.021/ 13.343/ 18.202    \cr  \hline 
     \centering Park and Kim (P) \cite{flow_As_2017MyP22} (0.05/0.1/0.15) &  \centering  52.085/ 26.005/ 17.456 &  \centering 10.082/ 11.456/ 16.619  \cr \hline 
     \centering Park and Kim (A) \cite{flow_As_2017MyP22} (0.05/0.1/0.15) &   \centering  68.914/ 34.558/ 22.946 &  \centering 13.224/ 14.554/ 19.437    \cr 
       \Xhline{3\arrayrulewidth} 
    \end{tabular}
\end{table}
\end{center} \footnotetext[5]{Simulation performed with 6,676,360 sample trajectories.}

\vspace{-0.8cm} 

\section{Conclusions} 
This paper proposes a conflict prediction technique for 2-D scenarios where vehicle motion can be described as a Gaussian process and the conflict boundary can be approximated by a number of straight-line segments. The proposed technique first reduces the dimensionality of the problem, and then writes the expression for the probability of conflict in terms of the FPTD. The final expression involves the computation of a single numerical integral, which results in significant computational savings when compared to a known fast method like probability flow. Simulation experiments were used to demonstrate the computational efficiency of the proposed method, as well demonstrating that the accuracy of the proposed method is of the same order as the accuracy of existing methods. Two variants of probability flow were used for comparison, as well as a recent ICP method. Increasing the number of line segments by a factor of $N$ would increase the computational time by a factor of $N$. However, since these calculations are completely independent of each other they can be computed in parallel. 



\bibliographystyle{unsrt} 
\bibliography{automatica_latex_Feb2023_ACCEPTED}           



\end{document}